\documentclass[journal]{IEEEtran}
\IEEEoverridecommandlockouts
\usepackage{graphicx} 
\usepackage{amsmath}
\usepackage{makecell}
\usepackage{graphicx}
\usepackage{textcomp}
\usepackage{stfloats}
\usepackage{array} 
\usepackage{longtable}
\usepackage{booktabs}
\usepackage{float}
\usepackage{subcaption}
\usepackage{booktabs,ragged2e}

\usepackage[flushleft]{threeparttable}
\usepackage{amssymb}
\usepackage{multirow}
\newcommand{\multirowoffset}{-0.5\dimexpr \aboverulesep + \belowrulesep + \cmidrulewidth}

\usepackage{color}
\usepackage{etoolbox}
\usepackage{pdfcomment}
\usepackage[normalem]{ulem}

\newtheorem{problem}{Problem}

\newtheorem{definition}{Definition}
\DeclareMathOperator{\E}{\mathbb{E}}

\usepackage{cite}
\usepackage{amsmath,amssymb,amsfonts}
\usepackage{algorithmic}
\usepackage{graphicx}
\usepackage{textcomp}
\usepackage{xcolor}
\def\BibTeX{{\rm B\kern-.05em{\sc i\kern-.025em b}\kern-.08em
    T\kern-.1667em\lower.7ex\hbox{E}\kern-.125emX}}

\title{FedVAE: Trajectory privacy preserving based on Federated Variational AutoEncoder
}

\author{
    \IEEEauthorblockN{Yuchen Jiang$^{1}$, Ying Wu$^{1}$, Shiyao Zhang$^{2}$, and James J.Q. Yu$^{3}$}
    \\
    \IEEEauthorblockA{$^{1}$Department of Computer Science and Engineering, Southern University of Science and Technology, China \\ $^{2}$Research Institute for Trustworthy Autonomous Systems, Southern University of Science and Technology, China \\
    $^{3}$Department of Computer Science, University of York, United Kingdom}\\
    \IEEEauthorblockA{E-mails: \{12232418, 12059004\}@mail.sustech.edu.cn, zhangsy@sustech.edu.cn, james.yu@york.ac.uk}

\thanks{This work is supported by the Stable Support Plan Program of Shenzhen Natural Science Fund No. 20220815111111002.} \thanks{Corresponding author: Shiyao Zhang and James J.Q. Yu.}
}
    
\begin{document}

\maketitle

\thispagestyle{empty}
\pagestyle{empty}

\begin{abstract}
    The use of trajectory data with abundant spatial-temporal information is pivotal in Intelligent Transport Systems (ITS) and various traffic system tasks. Location-Based Services (LBS) capitalize on this trajectory data to offer users personalized services tailored to their location information. However, this trajectory data contains sensitive information about users' movement patterns and habits, necessitating confidentiality and protection from unknown collectors. To address this challenge, privacy-preserving methods like K-anonymity and Differential Privacy have been proposed to safeguard private information in the dataset. Despite their effectiveness, these methods can impact the original features by introducing perturbations or generating unrealistic trajectory data, leading to suboptimal performance in downstream tasks. To overcome these limitations, we propose a Federated Variational AutoEncoder (FedVAE) approach, which effectively generates a new trajectory dataset while preserving the confidentiality of private information and retaining the structure of the original features. In addition, FedVAE leverages Variational AutoEncoder (VAE) to maintain the original feature space and generate new trajectory data, and incorporates Federated Learning (FL) during the training stage, ensuring that users' data remains locally stored to protect their personal information. The results demonstrate its superior performance compared to other existing methods, affirming FedVAE as a promising solution for enhancing data privacy and utility in location-based applications.
\end{abstract}


\section{Introduction}

With the increasing popularity of GPS-embedded devices, data collection, and data mining technologies, a vast amount of trajectory data has been generated \cite{wang2020survey}. This data is widely utilized in various domains, including travel time prediction \cite{zhu2022cross} and location-based services \cite{chekol2022survey}. These applications serve as crucial components in enabling the adaptive development of ITS. However, the collection of trajectory data raises significant privacy concerns as it may contain sensitive information about users' daily lives \cite{9772978}. If the data is not adequately protected, it can be susceptible to misuse and unauthorized access, potentially leading to privacy breaches and harm to individuals. Therefore, it is crucial to prioritize the protection of users' privacy when utilizing such data.

Recently, there has been a growing interest in the development of privacy-preserving methods for GPS trajectory data. \cite{chen2018mix} propose a method based on the K-Anonymity technique to generate new trajectory data from real datasets. Additionally, Differential Privacy (DP), introduced by \cite{dwork2009differential}, has been utilized to protect against possible attacks by applying Laplace Noise to location data. However, these methods often modify trajectory data, which could introduce bias and compromise the integrity of the entire database. Therefore, it is important to balance privacy preservation and data utility. To achieve this, we turn to machine learning methods that are efficient in extracting the features of the dataset. It is assumed that the dataset is generated from underlying latent space so that models can capture the latent representation of the dataset. Therefore, we need a generative model that approximates the data probability distribution by Bayesian inference as well as generating an entirely new dataset. Thus, the Variational AutoEncoder (VAE) model is proposed to generate a new dataset with a similar feature space and data distribution compared with the original dataset. Besides, Federated learning (FL) provides a way to maintain user privacy in centralized data settings for deep learning models. The training process occurs locally on devices or edge servers, allowing the model to learn from decentralized data without sending individual data to a central server. This approach strikes a balance between utilizing valuable data for model training and preserving user privacy, making it a promising framework for privacy-preserving deep learning in real-world applications \cite{9632695}.

\begin{figure}[ht]
\centering
\includegraphics[scale=0.4]{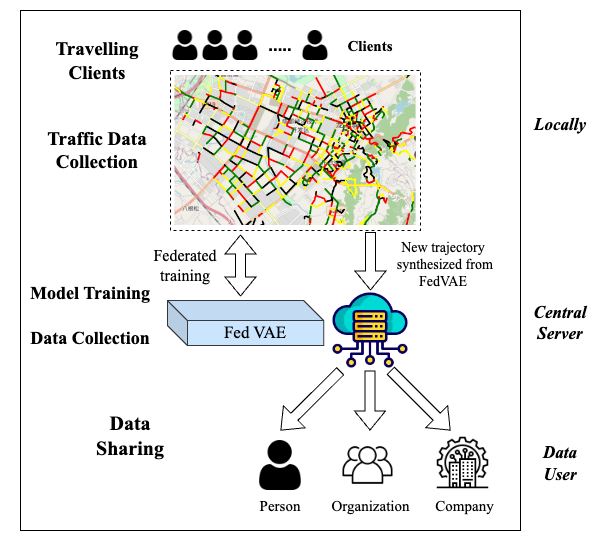}
\caption{The workflow of the proposed theme.}
\label{fig.wholeStructure}
\end{figure}

In this paper, we address the privacy-preserving task for trajectory datasets by employing the federated learning approach alongside trajectory generation methods. To contextualize our research, we comprehensively review and compare existing privacy-preserving methods specifically tailored for trajectory datasets. We also explore various trajectory data generation approaches and delve into the realm of federated learning, providing a crucial foundation for the subsequent methodology introduction. We take advantage of the FL framework and VAE model to propose a novel methodology, Federated Variational AutoEncoder (FedVAE), for generating new trajectory data based on actual data while preserving user privacy. We evaluate our approach in both privacy and utility metrics. Especially, our approach is evaluated in a practical downstream task, namely, traffic mode identification (TMI), to demonstrate its utility. The experimental results showcase the exceptional data utility of the generated dataset while maintaining user location privacy. The workflow of the proposed theme is illustrated in Figure \ref{fig.wholeStructure}. By federated learning, local clients participate in VAE training and synthesize new trajectory datasets from well-trained VAE that can be safely used by the central server and shared with data users. Highlights of this paper are summarized as follows:

\begin{itemize}
\item Our work represents the first attempt to apply the FL training framework to the trajectory generation problem, incorporating considerations for both privacy preservation and maintaining the original feature space.

\item FedVAE offers a decentralized learning and generative method, which not only keeps original data locally but also provides a completely new dataset with high utility and low risk of privacy leakage for subsequent uses.

\item The proposed approach can be extended to other privacy-preserving downstream tasks in ITS. Our work contributes to the field by addressing the privacy concerns associated with trajectory data generation, opening avenues for further research in privacy-preserving machine learning techniques.
\end{itemize}

The remainder of this paper is organized as follows. Section \ref{sec:relatedwork} introduces the related work on trajectory privacy preservation, trajectory generation, and federated learning. Sections \ref{sec:preliminary} and \ref{sec:methodology} present the problem definition and the proposed method. Section \ref{sec:experiment} and \ref{sec:discussion} show the experimental setup and analysis together with discussions about the proposed model, while Section \ref{sec:conclusion} concludes the paper and provides directions for future research.

\section{Related Work} \label{sec:relatedwork}

\subsection{Trajectory Privacy preserving}
Over the past few decades, researchers have proposed various approaches to address the challenge of preserving location privacy in LBS. These approaches can be broadly classified into two main architectures: centralized and non-centralized.

In a centralized architecture, a centralized entity is employed to protect location privacy. The most widely used method is K-anonymity \cite{sweeney2002k}, which ensures that each data point is indistinguishable from at least k-1 other data points \cite{terro2008privacy, dai2015segment}. It is commonly applied in sensitive databases to prevent the identification of unique users through queries. On the other hand, non-centralized architectures include obfuscation-based methods, such as introducing perturbations into collected or quantizing locations \cite{dwork2009differential}, as well as cryptographic-based methods \cite{9339974}. However, these approaches have their drawbacks. For example, finding a trusted third party for centralized methods and cryptographic-based methods \cite{9339974} is a challenging task. Besides, obfuscation-based methods can be vulnerable to background knowledge attacks \cite{de2019give}, where attackers can apply additional information to de-anonymize users. 

To summarize, previous methods for preserving privacy in trajectory data overlooked semantic information, resulting in a loss of dataset utility. Additionally, approaches involving cryptographic or clustering operations are unsuitable for handling sparse and large datasets. Hence, a trajectory generation-based method is proposed using a VAE model, maintaining the original feature space and producing new data. In line with real-life scenarios, FL is applied to the training process, ensuring users' data is kept locally and not exposed to other devices, thus protecting users' privacy and improving operational efficiency.

\subsection{Trajectory generation}
Trajectory generation methods are utilized to enhance the feature space in sparse data or reduce noise in noisy datasets, making the data more informative and useful for subsequent tasks. There are several methods for generating trajectory data. \cite{sun2023synthesizing} generates new trajectory data by space discretization and synthesis according to mobility patterns and movable constraints. \cite{rao2020lstm, demetriou2020generation} use machine learning to generate a new dataset from the original dataset. However, they are unsuitable for our goal of maintaining the original feature space, which is important in subsequent tasks. Besides, training GAN-based models can be challenging due to gradient vanishing and explosion issues. Therefore, we turn to VAE-based models, generally used for data augmentation \cite{liu2022vaeaug}. They focus on both reconstruction and distribution loss, making them effective for data generation.

\subsection{Federated Learning}

In the conventional machine learning paradigm, data is typically sent to central servers for training. However, this raises concerns about privacy and data security. To address these concerns, FL has emerged as a solution to facilitate communication between local clients and central servers without transferring the original data \cite{9599369}. The key idea is that the clients contribute to the model training process by uploading their local updates of the model's gradients to the servers instead of sharing raw data. Then these updates are aggregated on the servers to create a comprehensive model. This approach ensures the privacy and security of the data, as the raw data remains on the client's local devices. 

In the context of FL, a commonly used approach is federated averaging (FedAVG) \cite{mcmahan2017communication}. In FedAVG, the locally trained models are weighted according to the size of the respective local dataset. During each communication round, the clients' weight updates of their local models are transmitted to the central server. The server then averages these updates and sends the resulting average back to the federation of clients. This averaged value is then used to initialize the next round of the training process. In our work, we deployed FedAVG as our fundamental training framework due to its simplicity and communication-efficiency advantages, particularly when training on mobile and edge devices in trajectory data applications. Considering real-life scenarios, communication overhead needs to be viewed during federated learning \cite{9609654}. Joint-Announcement Protocol \cite{9082655} is proposed to handle the large-scale scenario where the FedAVG algorithm is hard to converge because of expensive communication overhead.

\section{preliminary}\label{sec:preliminary}

This section provides several preliminary definitions related to our research target: GPS trajectory and the concepts of federated learning, including local clients and the central server. Additionally, we define our research problem as the generation of trajectories while preserving privacy and addressing one downstream problem of travel model identification.

\begin{definition}[GPS Trajectory]\label{defTraj}

    Let $\mathcal{P}$ denote a sequence of consecutively sampled GPS points, i.e., $\mathcal{P}=\{p_1,\dots,p_m\}$, where $p_i=[\text{lat}_{i}, \text{lon}_{i}, t_i]$, $i \in \{1,\dots,m\}$, represents the latitude and longitude of the GPS point at the time step $t_i$, indicating the device's (or the user's) location at that time.

\end{definition}

\begin{definition}[Local Clients]

    Let $\mathcal{C} = \begin{Bmatrix}C_{1}, C_{2}, ..., C_{k}\end{Bmatrix}$ denote a set of local clients (i.e., individual mobile devices) with $k$ groups. Each local client $C_{i}$ possesses a distinct set of GPS trajectories denoted as $\mathcal{D} = \begin{Bmatrix}\mathcal{D}_{1}, \mathcal{D}_{2}, ..., \mathcal{D}_{k}\end{Bmatrix}$, which satisfied $ \mathcal{D}_{i} \cap \mathcal{D}_{j} = \emptyset $. Organizations treat the collection of distributed models $\mathcal{F} = \begin{Bmatrix}F_{1}, F_{2}, ..., F_{k}\end{Bmatrix}$ as local models. These models are trained independently and simultaneously, utilizing the training data stored in the respective GPS trajectory sets $\mathcal{D}$.

\end{definition} 

\begin{definition}[Central Server]
    The central server, serving as a third-party entity, resides in the cloud infrastructure. Its primary function is to aggregate the local models $\mathcal{F} $ contributed by individual clients in order to generate a global model $\hat{F}$. Subsequently, the central server disseminates this global model to all or part of participating local clients. 
\end{definition} 

\begin{problem}[Trajectory Generation]
Given a sequence of consecutively GPS points $\mathcal{P}$, generate a new trajectory $\mathcal{P}'$ that is similar to $\mathcal{P}$ but does not reveal the user's identity, which is formulated as $\mathcal{P} \overset{f(\cdot)}{\rightarrow} \mathcal{P}'$.
\end{problem}

\begin{problem}[Travel Model Identification]
Given a sequence of GPS points $\mathcal{P}_m$, we can then reform the GPS segment where $\text{label}_m$ represents the mode of transportation for segment $k$. We aim to train a function $g(\cdot)$ to identify the mode of transportation for each trajectory segment, which is formulated as $\mathcal{P}_m \overset{g(\cdot)}{\rightarrow} \text{label}_m$.
\end{problem}

In summary, the goal of this paper is to generate a new trajectory $\mathcal{P}'$ that is similar to $\mathcal{P}$ but does not reveal the user's identity in each client $C_{k}$. The gradient of the individual client's model $f_{k}$ is uploaded to a central server for aggregation. The central server then updates its global model $\hat{F}$ based on the aggregated gradients. The updated $\hat{F}$ is then disseminated to all or part of the participating local clients $\mathcal{F}$ for further training. The training process iterates until convergence of the global model is achieved, ensuring that further improvements are made through multiple iterations.

\section{Methodology}\label{sec:methodology}

In this section, we introduce the proposed VAE model for trajectory generation within the local client, which describes the mechanics of VAE. We also present the federated learning framework, which enables the training process, thereby achieving communication-efficient training.

\subsection{Local VAE Model for Trajectory Generation}
\label{sec:fedvae}

\begin{figure}[h]
    \centering
    \includegraphics[scale=0.235]{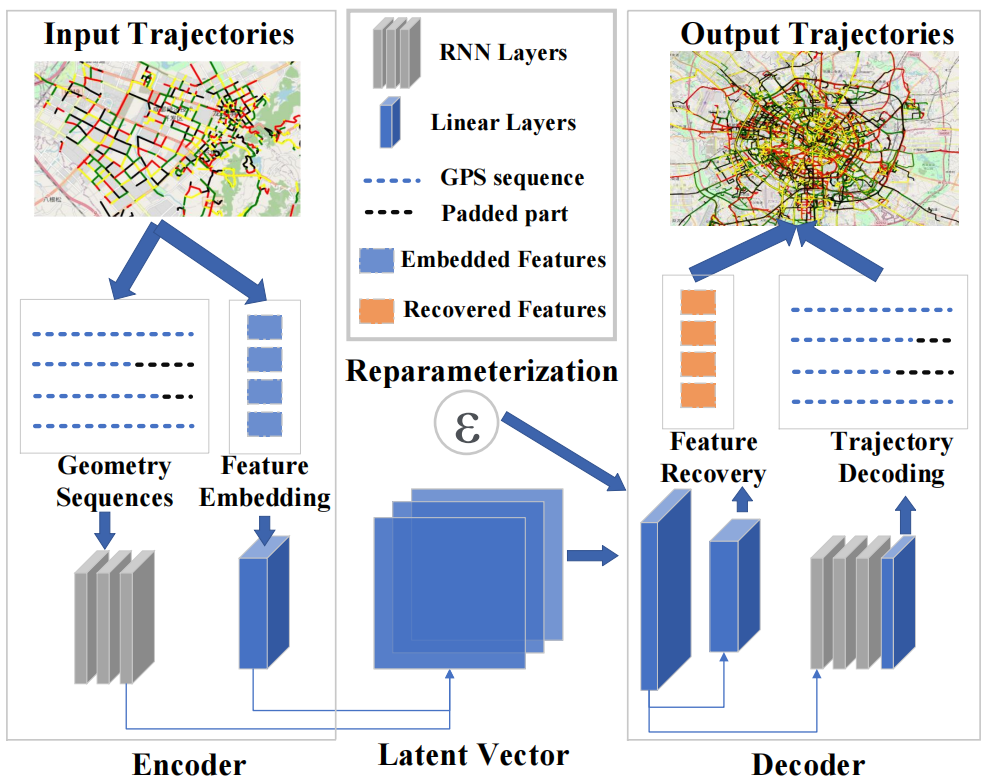}
    \caption{Local VAE model for trajectory generation.}
    \label{fig.fedvae}
    \end{figure}

We propose a VAE model for generating new trajectory data in the local clients, as illustrated in Figure \ref{fig.fedvae}.  It consists of two fundamental components: an encoder and a decoder. The encoder module aims to capture the underlying latent representation of the input data (i.e., GPS trajectory and travel mode labels). In contrast, the decoder module focuses on generating synthetic data that aligns with the distribution of the original training data from the latent vector. 

\subsubsection{Encoder module}
\label{subsec: encoder}

Given a collection of GPS trajectories, we first pass them into Recurrent Neural Network (RNN)-based layers to capture the sequential features. Additionally, linear layers are employed to incorporate auxiliary features such as traffic modes and related attributes into the model. The resulting encoder produces a latent representation $z$, which is further transformed via a non-linear function to produce a probability distribution $p_{\theta}(x)$ over the latent space. Subsequently, a set of generative latent vectors is sampled from the distribution and passed into the decoder module to generate synthetic trajectory data. 

However, directly computing the probability distribution $p_{\theta}(x)$ is computationally challenging due to its complexity. To overcome this issue, Bayesian inference is employed by incorporating a prior distribution $p_{\theta}(z)$ and the true posterior distribution $p_{\theta}(x|z)$ \cite{David2017Variational}. In this paper, we assume a standard Gaussian distribution as the prior distribution $p_{\theta}(z)$. The true posterior distribution is then approximated using variational inference \cite{Liang2018Variational, David2017Variational} by minimizing the Kullback-Leibler (KL) divergence between $p_{\theta}(x|z)$ and an approximate posterior distribution $q_{\phi}(x|z)$. Instead of directly using a variational distribution, we employ an inference model parameterized by $\phi$ to generate the mean $\mu$ and variance $\sigma$ of the approximating variational distribution. This allows us to formulate the minimization process as follows:

\begin{equation}
    \begin{aligned}
    \ln p_{\theta}(x) = L_{\theta, \phi}(x) + \mathrm{KL}(q_{\phi}(z|x)||p_{\theta}(z|x)),
    \end{aligned}
    \label{eq:2}
    \end{equation}
where $\ln(\cdot)$ is the natural logarithm operator. 

This form can be obtained by averaging the objective function over all data as:
\begin{equation}
    \begin{aligned}
    L_{\theta, \phi}(x) = {\E}_{q_{\phi}(z|x)}[\ln p_{\theta}(x|z)] - \mathrm{KL}(q_{\phi}(z|x)||p_{\theta}(z)).
    \end{aligned}
    \label{eq:3}
\end{equation}

The first term in Eq. \ref{eq:3} corresponds to the reconstruction loss, which quantifies the negative log-likelihood of the input data $x$ given the latent vector $z$. It measures how well the decoder reconstructs the original input. The second term represents the regularization term, which computes the Kullback-Leibler (KL) divergence between the posterior distribution $q_{\phi}(z|x)$ and the prior distribution $p_{\theta}(z)$. This regularization term encourages the posterior distribution to closely match the prior distribution closely, thereby preventing the model from overfitting and ensuring a more structured latent space.

We can obtain an unbiased estimation of Eq. \ref{eq:3} by sampling $z \sim q_{\phi}$ and performing stochastic gradient ascent to optimize it. However, a challenge arises when updating the model's weights due to the inability to differentiate the sampling operation from a Gaussian distribution during backpropagation. To address this, we employ the reparameterization trick \cite{Stochastic2014Rezende}. It involves sampling $\epsilon$ from a standard Gaussian distribution $\epsilon \sim N(0, I)$ and constructing the latent vector as $z = \mu + \sigma \epsilon$.

\subsubsection{Decoder module}

After sampling from the posterior distribution $z \sim q_{\phi}$, the decoder module is responsible for reconstructing the original input data $x$ from the latent vector $z$. The decoder module is composed of RNN-based layers and linear layers. The RNN-based layers and the linear layer decode the latent vector $z$ into trajectory decoding $\hat{\text{Tr}_i}$ and feature recovery $\hat{\text{Em}_i}$ respectively, which are subsequently fed into the linear layers to reconstruct the original input data.

\subsubsection{Optimization}

Our loss function comprises two components: the reconstruction loss $L_{r}$  and the Kullback-Leibler (KL) divergence loss $L_{kl}$, as mentioned in Section \ref{subsec: encoder}. The KL divergence loss ensures the alignment between the latent vector space and a standard Gaussian distribution, regulating the distribution difference. The reconstruction loss consists of two parts: the trajectory generation loss and the embedding feature loss. Our paper focuses on the downstream task of traffic mode identification. Hence, the trajectory generation loss is defined as the Mean Squared Error (MSE) between the trajectory decoding $\hat{\text{Tr}_i}$ and the original trajectory $\text{Tr}_i$. Meanwhile, the embedding feature loss is calculated as the Cross-Entropy Loss (CEL) between the embedding feature $\text{Em}_i$ and the feature recovery $\hat{\text{Em}_i}$. The optimization target $L$ of the proposed model is formulated as follows:

\begin{equation}
    \begin{aligned}
        L &= L_r + L_{kl} \\
        &= \frac{1}{n} \sum_{i=1}^{n} (\text{Tr}_i - \hat{\text{Tr}}_i)^2-\sum_{i=1}^{n} \text{Em}_i \log(\hat{\text{Em}}_i) \\
        &+ \mathrm{KL}(q_{\phi}(z|x)||p_{\theta}(z)),\\
    \end{aligned}
\end{equation}
where $\log(\cdot)$ is the logarithm operator.

We employ a truncated function in our trajectory generation loss to improve the generalization of newly generated trajectories. We establish a threshold to ensure that the reconstruction loss remains within a reasonable range for real-life scenarios. When the reconstruction loss surpasses this threshold, it is truncated to zero, resulting in only the KL divergence loss ($L_{kl}$) being utilized for model updates. In this study, we set the threshold to 0.000625 for MSE reconstruction loss, corresponding to a geographic distance error of fewer than 4 kilometers for the reconstructed trajectories.

\subsection{Global Federated Training Framework}

\begin{figure}[h]
\centering
\includegraphics[scale=0.35]{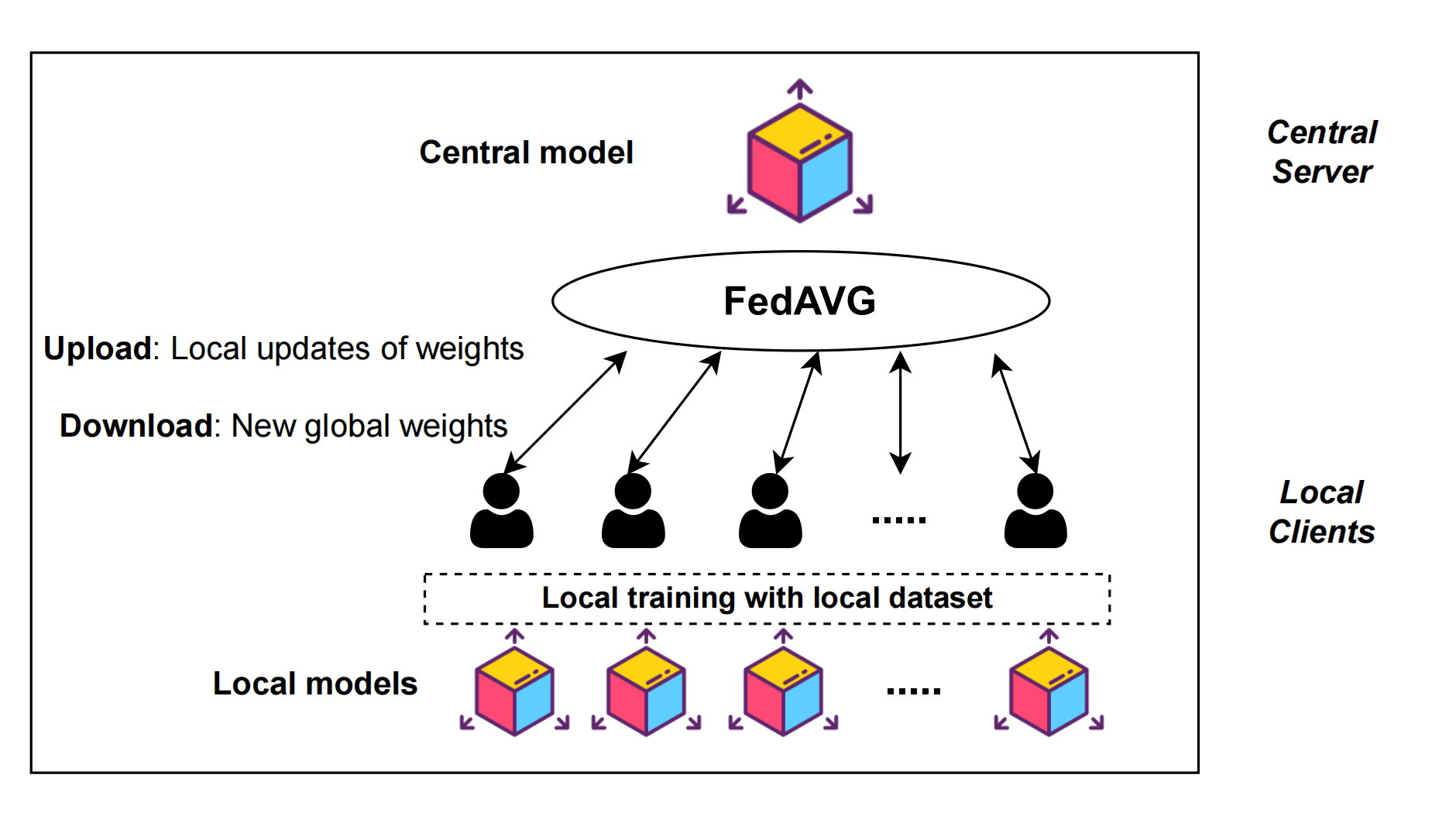}
\caption{Federated training theme.}
\label{fig.fltheme}
\end{figure}
Figure \ref{fig.fltheme} shows the whole training framework. Details are described below:
\subsubsection{Initialization}
The central server has an initial central model and each local client has an initial local model. During each epoch, the model of each client is initialized with the weights of the central model.

\subsubsection{Training}
The training process is distributed across client servers instead of central servers. As described above, to minimize the amount of raw data stored on central servers, clients contribute to the model training process by uploading their local updates of the model's gradients. 

\subsubsection{Aggregation}
Subsequently, the central server uses the FedAVG algorithm to update the model's weights from clients' gradients. The federated framework ensures that the original data is not directly collected by the central server while still enabling the successful training of the global model necessary for our specific objectives. 

\subsubsection{Communication overhead}
Considering realistic situations, communication overhead needs to be discussed. When the client number is small, a small-scale federated learning model is built up by the FedAVG algorithm. When large-scale scenarios occur with a large number of clients, FedAVG is hard to converge due to expensive communication overhead. Thus, Joint-Announcement protocol \cite{9082655} is applied. When facing large-scale scenarios, local clients will check in with cloud servers to ensure connectivity before training. The central server randomly selects a certain percentage of clients in the $i$-th round training. The protocol avoids failures between clients and the central server, reducing communication overhead and execution time in federated training.

\section{Experiment}\label{sec:experiment}
In this section, we introduce the dataset, competing baseline, experiment set, and evaluation metrics in detail. We analyze the results in terms of privacy analysis with anonymization, trajectory statistics between the original trajectory data and the generated data, as well as the utility application in the downstream task TMI.

\subsection{Dataset}
We employ Microsoft's Geolife dataset \cite{zheng2008learning, zheng2008understanding} for our experiment, which is extensively utilized for traffic mode identification \cite{zhu2021improving}. The dataset comprises both labeled and unlabeled data, collected from a total of 182 users, including 69 users with labeled data and 113 users with unlabeled data. For five years, the users record their traffic mode data in the form of Global Positioning System (GPS) points and timestamps, along with the corresponding traffic mode, which encompasses activities such as walking, biking, and bus transit. 

Moreover, the collection of raw data from users varies in length, thus requiring pre-processing steps such as cutting and padding. According to the Geolife dataset, we divide each trajectory into sub-sequences of length 100, which is a reasonable length that contains enough traffic mode information, and pads shorter sequences with zeros. Additionally, a mask matrix is constructed to record padded segments and distinguish valid and invalid parts for loss computation by binary values. For unpadded segments, the mask is a sequence of ones with a length of 100.

\subsection{Competing Baseline} 
To demonstrate the applicability of FedVAE, we adopt three representative methods for traffic mode identification, which are K-Neighbor and MLP and Semi-supervised Convolutional AutoEncoder. We also select three privacy-preserving methods for trajectory data and compare their performance in the traffic mode identification task. It is noticed that the performance is evaluated by accuracy and variance in TMI tasks.

\begin{itemize}
    \item Perturbation. This approach randomly perturbs each location point so that the privacy information will be ambiguous. 
    \item Mixzone \cite{chen2018mix}. Mixzone aims to find a spatiotemporal zone in which multiple trajectories traverse. Then mixing and perturbing operations are applied to these trajectory data so that the adversaries can not identify each trajectory.
    \item K-Anonymity \cite{dai2015segment}. This method generates another $k-1$ trajectory based on K-Anonymity. It generates $k-1$ points around each selected sensitive point with private information like homes and workplaces. It randomly connects these $k-1$ location point pairs with other nonsensitive location points into $k-1$ trajectories to protect the original trajectory.
\end{itemize}

Besides, we controlled the hyper-parameters of these baseline methods to get the optimal performance. From the given range of parameters, we test their performances in the TMI task to judge the best parameter combination. For Mixzone, we choose the best pair of hyper-parameters with k as 6, which stands for the number of users passing through the zone, and L-limit as 0.1, which controls the length limit of each trajectory included in the zone. As for K-Anonymity, we choose k as 5, which represents the trajectory anonymous degree, to achieve optimal performance.

\subsection{Experiment settings} 
We implemented our model using PyTorch. The model is trained on the server with 2.1 GHz Intel Xeon CPU E5-2620 v4 CPUs, 128G RAM, and nVidia 2080Ti 11G GPUs. The model is trained using Adam optimizer with an initial learning rate of 0.01, and the learning rate turns to 0.001 when 500 epochs are finished. The batch size is set to 2000. We split the whole dataset into train, test and validation sets with the rate of 8:1:1 to avoid over-fitting. The model is trained for 2000 epochs, and early stopping is applied when the loss is not changing anymore. 

Besides, for the federated training stage, we set the client number as 69 to meet the real-world situation, as our dataset includes 69 users with labeled data. FedAVG algorithm together with the Joint-Announcement protocol is used to handle large-scale training scenarios. The hyperparameters in the model are the same for each local client.

\subsection{Evaluation Metrics}
\begin{table*}[]
     \caption{Comparison of privacy-preserving methods for TMI task performance. (Similarity represents the privacy protection metric ($\downarrow$). Utility is measured by accuracy ($\uparrow$) and variance ($\downarrow$)). }
     \label{tab:evaluation}
     \centering
     \begin{tabular}{llllllll}
     \toprule
     \multirow{2}{*}[\multirowoffset]{\textbf{Metric}}  & \multirow{2}{*}[\multirowoffset]{\textbf{Base model}} & \multicolumn{5}{c}{\textbf{Privacy-preserving model}}                            &  \\ \cmidrule{3-8} 
                              &                             & Baseline     & Perturbation & MixZone      & K-Anonymity  & \textbf{FedVAE (ours)} &  \textbf{Improved by}       \\ \midrule
     \textbf{Similarity}               &                             & -            & 0.81         & 0.51         & 0.74         & \textbf{0.20}         &  \textbf{60.78\%}  \\
     \multirow{3}{*}{\textbf{Utility}} & K-Neighbor                  & 57.87\% ± 0.01 & 70.03\% ± 0.02 & 56.98\% ± 0.01 & 58.49\% ± 0.01 & \textbf{82.53\% ± 0.02}  &  \textbf{17.85\%}    \\
                              & MLP                         & 47.98\% ± 0.07 & 64.26\% ± 0.08 & 41.26\% ± 0.10 & 43.96\% ± 0.05 & \textbf{78.35\% ± 0.02}  &    \textbf{21.93\%}     \\
                              & Semi-Supervised             & 73.28\% ± 0.01 & 83.82\% ± 0.02  & 76.09\% ± 0.01 & 81.36\% ± 0.02 & \textbf{91.04\% ± 0.01}  &   \textbf{8.61\%}   \\ \bottomrule
     \end{tabular}
\end{table*}

We employ three metrics to evaluate the performance of the proposed models. Firstly, privacy protection performance is evaluated in terms of anonymized level, by comparing similarity across different travel modes. Lower similarity indicates a higher anonymized level. Additionally, we evaluate the statistical properties of the generated and original trajectory distributions using KL divergence. A lower KL divergence signifies a more similar distribution of trajectories between them. Furthermore, we compare the experiment results and show the improvement between our and the best of others.

\textbf{Privacy metric:} The dataset's privacy level is evaluated by the similarity of the original and generated one. Higher similarity means a higher risk of attacks as attackers can identify the user of the trajectories when generated trajectories are similar to the original one. We evaluate the similarity by applying relative distance between GPS points on the trajectory. Given two trajectories $\mathcal{P}$ and $\mathcal{P}'$ and their traffic mode, we define their similarity of them:
\begin{equation}
    \begin{aligned}
     \mathrm{Similarity}(\mathcal{P}, \mathcal{P}', mode) &= \alpha\dfrac{\beta_{mode}}{\frac{1}{n}dist(\mathcal{P}, \mathcal{P}')} \\
    & = \alpha\dfrac{\beta_{mode}}{\frac{1}{n}\sum_{i=1}^{n} dist(p_i, p'_i)}
    \end{aligned}
    \label{eq:eva_privacy}
    \end{equation}

 where $dist$ computes the geographic distance between two GPS points in the trajectories. $\alpha$ is the proportionality coefficient, and $\beta_{mode}$ measures the varying scale of similarity based on different traffic modes. In this paper, $\mathrm{traffic mode} = \{\mathrm{bus}, \mathrm{car}, \mathrm{walking}, \mathrm{biking}, \mathrm{subway} \}$, and $\beta_{mode}$ is set accordingly $\{\mathrm{0.5 km}, \mathrm{1 km}, \mathrm{0.05 km}, \mathrm{0.2km}, \mathrm{3 km} \}$ for each mode. To simplify, $\alpha$ is set to 1. \footnote{It is noted that K-Anonymity generates additional trajectories instead of modifying original trajectories. Therefore, we compute the similarity between the original trajectory and other $k-1$ trajectories.}

\textbf{Trajectory statistics:} We investigate the trajectory distribution between the original and generated trajectory. We examine the similarities between the synthetic and original trajectories through 
$\mathrm{KL}$, a non-parametric method of comparing two one-dimensional distributions.

\textbf{Data utility:} We evaluate the utility of the generated dataset through its performance in downstream tasks. This paper focuses on the TMI task as our downstream task due to its prevalent application in real-world scenarios \cite{dabiri2018inferring, dabiri2020semisupervised}. Specifically, we employ classification accuracy and standard variance of the TMI task as metrics to evaluate the utility of the generated dataset. Higher accuracy and lower variance indicate better performance in the TMI task.

\subsection{Experiment Analysis}
Table \ref{tab:evaluation} provides a comprehensive comparison of various privacy-preserving methods utilized in the context of the TMI task. The evaluation revolves around two key metrics: \textit{Similarity}, which reflects the level of privacy protection (lower values indicate better privacy anonymization), and \textit{Utility}, which is assessed through classification accuracy (higher values are desirable) and variance (lower values are preferred) in the TMI task. The evaluation experiments are conducted several times and the \textbf{average} of results are recorded.

\textbf{Privacy Analysis:} 
As shown in Table \ref{tab:evaluation}, the proposed FedVAE model achieves the most substantial privacy preservation with a low \textit{Similarity} value of 0.20, outperforming other models. It demonstrates a remarkable 60.78\% enhancement in privacy protection. FedVAE generates synthetic trajectories with minimal similarity to the original data. By creating entirely new trajectories, the model reduces the risk of privacy attacks and protects individual user information effectively.

\textbf{Utility Analysis:}
The utility of the generated trajectory datasets is evaluated using the travel mode classification accuracy and its variance, as depicted in Table \ref{tab:evaluation}. The proposed FedVAE model demonstrates superior accuracy across all base models, achieving values of 82.53\%, 78.35\%, and 91.04\% for the K-Neighbor, MLP, and Semi-Supervised base models, respectively. These accuracy values indicate the FedVAE model's effectiveness in generating synthetic trajectories that preserve utility for various models in the TMI task. 

\begin{figure}[h]
    \centering
    \includegraphics[scale=0.147]{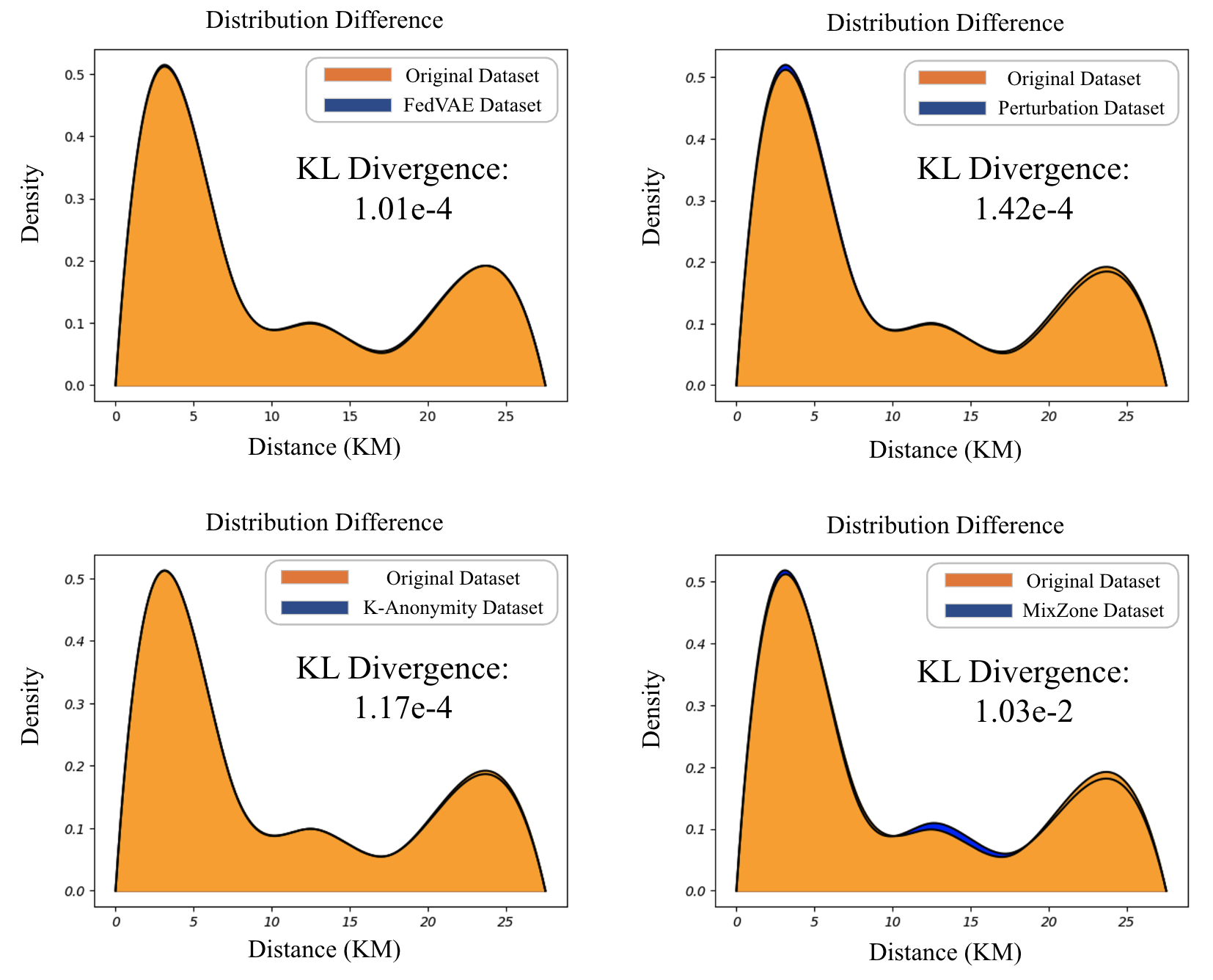}
    \caption{KL divergence comparison among four generated trajectory datasets and the original dataset.}
    \label{fig.klloss}
    \end{figure}
    
\textbf{Trajectory Statistics:} 
Figure \ref{fig.klloss} presents a comparison of the data distribution between the original dataset and the generated dataset. The color difference exhibited in the figure signifies variations in data distribution. Notably, the majority of trajectories in the dataset fall within the range of over $\mathrm{20 km}$ or under $\mathrm{10 km}$, owing to the specific traffic modes and real-life traffic scenarios considered. Across all methods, except for MixZone, the generated trajectories exhibit similar distribution patterns to the original dataset. The proposed FedVAE model successfully captures the essential characteristics of the original dataset, preserving trajectory distances in alignment with real-life traffic situations. In addition, it can capture the latent representation of the dataset and approximate the data probability distribution, leading to the lowest KL divergence between the generated dataset and the original dataset.

\section{discussion}\label{sec:discussion}
\subsubsection{Model Scalability}
In addition to the aforementioned experiments, we have also conducted tests on our method to evaluate its stability in the federated learning structure, which is crucial for ensuring its continued effectiveness as the number of clients increases or decreases. The model successfully converged regardless of the variation in the number of clients. The results are shown in Table \ref{tab:scalability}, which guarantees the scalability of the model.

\begin{table}[h]
    \caption{Scalability test.}
    \label{tab:scalability}
    \centering
    \begin{tabular}{cccccccc}
    \toprule  
    Client Number&2&5&10&20&30&50&70\\
    \cmidrule(lr){1-8} 
    Fitted Epochs&34&96&215&457&723&1264&1529\\
    \bottomrule 
    \end{tabular}
\end{table}

\subsubsection{Application}
Based on the statistics in Table \ref{tab:evaluation}, several methods achieve better accuracy than the baseline in the unique models. Since noises exist in the original dataset, data-enhanced methods can improve the robustness and utility of the dataset. Perturbation successfully achieves it, but it faces challenges in selecting appropriate noise scales and is vulnerable to attacks. Furthermore, the MixZone method and the K-Anonymity method have the potential risk of decreased performance. Our method successfully learns the representation of the dataset and achieves the best results.

\section{Conclusion}\label{sec:conclusion}
In this research, we proposed a FedVAE model to address privacy-preserving tasks for trajectory datasets, which combines federated learning and VAE techniques to maintain the original feature space and ensure privacy protection when generating synthetic trajectories. Our comprehensive experiments and analysis show that FedVAE outperforms other privacy-preserving models, achieving the lowest similarity value which indicates its effectiveness in the privacy protection task, and the highest travel mode identification accuracy which exhibits great utility in a downstream task. Furthermore, the generated trajectories showed similar patterns when compared with the original trajectory data.

\bibliographystyle{IEEEtran}
\bibliography{IEEEabrv, myrefs}

\addtolength{\textheight}{-12cm}   



\end{document}